\documentclass{Interspeech}
\usepackage{times}
\usepackage{latexsym}
\usepackage{comment}
\usepackage{xcolor}
\usepackage{hyperref}
\usepackage[T1]{fontenc}

\usepackage[utf8]{inputenc}

\usepackage{microtype}

\usepackage{inconsolata}

\usepackage{graphicx}

\usepackage{amssymb}
\usepackage{amsfonts}
\usepackage{amsmath}

\usepackage{booktabs}
\usepackage{lettrine}
\usepackage{makecell}

\usepackage{caption}     
\usepackage{tabularx}    
\usepackage{multirow}
\usepackage{siunitx}     
\usepackage{setspace}
\usepackage{subcaption} 

\interspeechcameraready

\title{Enhancing Speech Instruction Understanding and Disambiguation in Robotics via Speech Prosody}

\author[affiliation={1}]{David}{Sasu}
\author[affiliation={2}]{Kweku}{Andoh Yamoah}
\author[affiliation={3}]{Benedict}{Quartey}
\author[affiliation={1,*}]{Natalie}{Schluter}

\affiliation{Computer Science}{IT University of Copenhagen}{Denmark}
\affiliation{Intelligent Agents Research Group}{University of Florida}{U.S.A}
\affiliation{Intelligent Robot Lab}{Brown University}{U.S.A}

\email{dasa@itu.dk, kyamoah@ufl.edu, benedict\_quartey@brown.edu, nael@itu.dk}
\keywords{speech recognition, human-computer interaction, computational paralinguistics, speech prosody, robot planning, speech understanding, speech disambiguation}

\usepackage{comment}

\newcolumntype{C}{>{\centering\arraybackslash}X}
\newcolumntype{L}{>{\raggedright\arraybackslash}X}
\newcolumntype{R}{>{\raggedleft\arraybackslash}X}

\begin{document}

\maketitle

\begin{abstract}
    
    Enabling robots to accurately interpret and execute spoken language instructions is essential for effective human-robot collaboration. Traditional methods rely on speech recognition to transcribe speech into text, often discarding crucial prosodic cues needed for disambiguating intent. We propose a novel approach that directly leverages speech prosody to infer and resolve instruction intent. Predicted intents are integrated into large language models via in-context learning to disambiguate and select appropriate task plans. Additionally, we present the first ambiguous speech dataset for robotics, designed to advance research in speech disambiguation. Our method achieves 95.79\% accuracy in detecting referent intents within an utterance and determines the intended task plan of ambiguous instructions with 71.96\% accuracy, demonstrating its potential to significantly improve human-robot communication.
\end{abstract}

\section{Introduction}

\begin{figure}[t]
    \centering
    \includegraphics[width=0.5\textwidth]{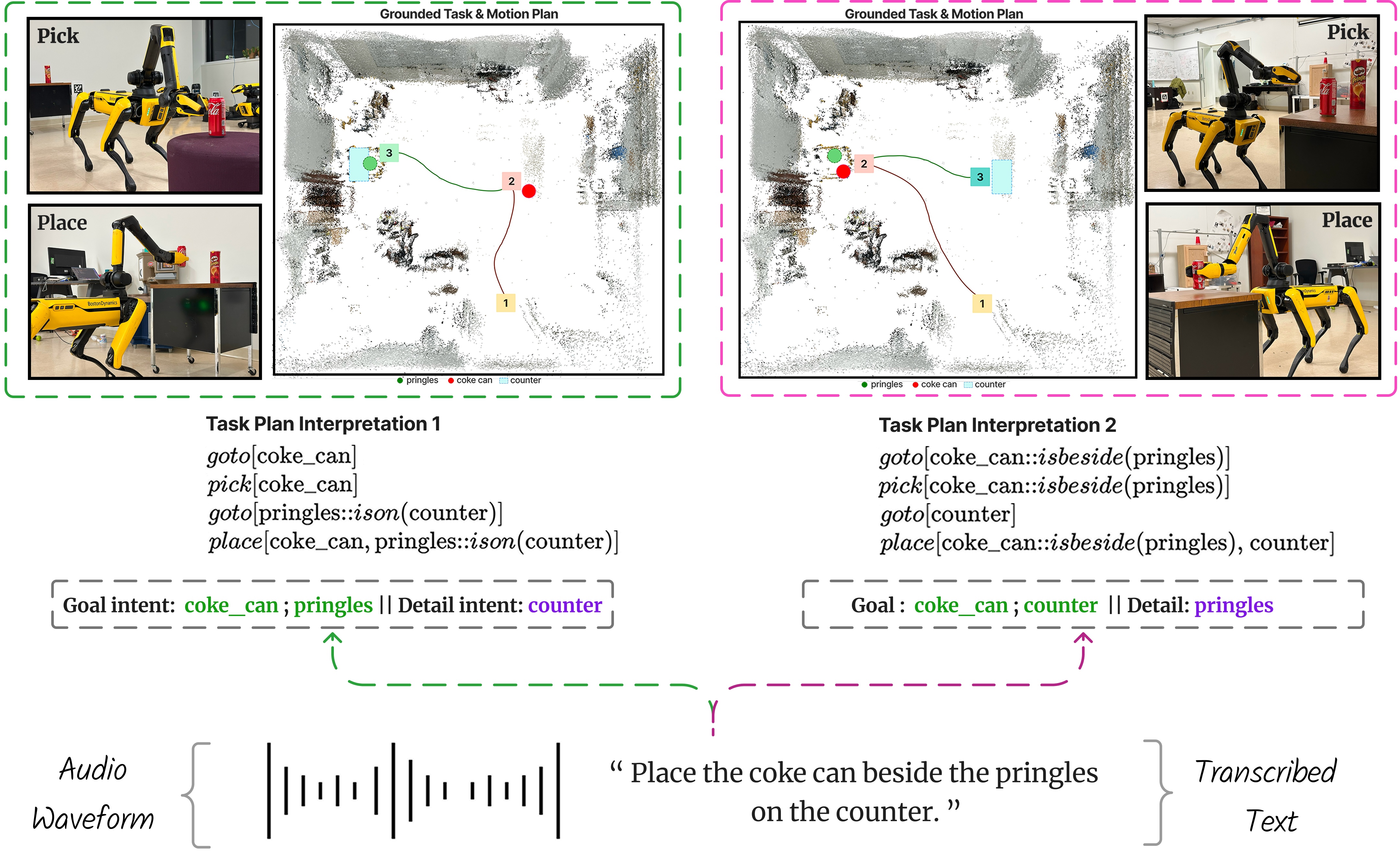}
\captionof{figure}{
Illustration of two distinct interpretations of the instruction
\emph{“Place the coke can beside the pringles on the counter.”}
\textbf{Left (green)}: The robot picks up the coke can and places it next to the pringles on the counter. 
\textbf{Right (pink)}: The robot locates the coke can already beside the pringles and relocates it onto the counter. 
Each plan is represented by a pointcloud of the scene and the robot’s motion path, while the robot images illustrate the pick-and-place actions. 
}
    \label{fig:1}
    \vspace{-0.5cm}
\end{figure}

Understanding and executing natural language instructions is a fundamental goal in human-robot interaction. To accurately follow instructions, robots must grasp several key elements such as the speaker’s intent, relevant referents, and execution constraints. However, instructions conveyed through text alone often fail to capture the wide variety of meanings that people intend. This limitation arises because rich speech prosodic cues—such as intonation, stress, and rhythm—that significantly affect the meaning of instructions are lost in textual representations. Consider the ambiguous instruction shown in Figure \ref{fig:1}: “Place the coke can beside the pringles on the counter”. One interpretation of this task is to place the coke can at a location beside the pringles, which happens to be on a counter. Another interpretation is to find the coke can (which is currently beside the pringles) and move it to the counter. Disambiguating such instructions relies on prosodic cues like pauses and emphasis, which convey the intended meaning. \looseness=-1

\footnote{* Currently at Apple.}


Current speech to motion robot systems predominantly rely on automatic speech recognition (ASR) to transcribe speech into text before processing \cite{tran.2023, davila2024voicecontrolinterfacesurgical, Rout2015SpeechRS}. While effective at capturing the literal words, ASR strips away essential prosodic cues like stress, rhythm, and intonation. This loss renders robots incapable of interpreting the speaker’s true intent in ambiguous or complex instructions. Existing approaches attempt to resolve ambiguity through contextual or semantic analysis of the transcribed text, but they fall short when critical information is embedded in the tonal aspects of speech \cite{McLoughlin_Indurkhya_2023, ai5030048}. \looseness=-1


To address this limitation, we propose a novel approach that explicitly leverages speech prosody to resolve ambiguities in spoken robot instructions. Concretely, we extract prosodic features from speech and employ an architecture that assigns an \textbf{intent label}---either \textbf{goal} (entities or locations to be acted upon) or \textbf{detail} (qualifiers specifying attributes or relationships)---to each word in an utterance. Reexamining the earlier example, in the first interpretation ``coke can'' and ``pringles'' would be classified as \textbf{goal intent referents}, while ``counter'' is a \textbf{detail intent referent} that qualifies the location of ``pringles''. In the alternative interpretation ``coke can'' and ``counter'' would be classified as \textbf{goal}, and ``pringles'' as a \textbf{detail} specifying the location of the coke can. We illustrate this distinction in interpretation and equivalent task plans in Figure \ref{fig:1}. We make three main contributions in this work:
\begin{enumerate}
    \item \textbf{Prosody-Aware Architecture.} We introduce an encoder-decoder model that exploits prosodic features for token-level intent classification, capturing long-range dependencies within spoken instructions. \looseness=-1
    \item \textbf{LLM Integration.} We demonstrate how prosody-driven intent predictions can be used with large language models, to select appropriate task plans from multiple candidates, bridging the gap between ambiguous human instructions and precise robot execution. \looseness=-1
    \item \textbf{Ambiguous Speech Dataset.} We highlight speech instruction disambiguation as a key research area in robotics and curate a novel dataset of \textbf{1,540} ambiguous utterances, to advance future work. \looseness=-1
\end{enumerate}

\section{Related works}
Recent works have demonstrated the effectiveness of incorporating prosody into speech understanding tasks such as intent classification \cite{noth2002use, shriberg2004prosody, rajaa2023improving, wei2022neural}. For instance, Wei et al. \cite{wei2022neural} proposed a neural prosody encoder that leverages prosodic information for end-to-end dialogue act classification, using a learnable gating mechanism to assess and selectively retain critical prosodic features. Other studies have explored prosody-based attention and distillation to improve end-to-end Spoken Language Understanding (SLU) \cite{rajaa2023improving}. While these approaches underscore the importance of prosodic features in SLU systems, they remain underexplored in human-robot interaction scenarios, where precise understanding of spoken instructions is essential for effective task execution. \looseness=-1

Early systems for robot instruction following from language often relied on hand-crafted grammars and symbolic representations to map natural language commands to executable actions \cite{macmahon2006walk}. Although these rule-based methods proved useful for navigation and simple tasks, their reliance on limited vocabularies and rigid syntactic structures restricted their ability to handle the nuances of spontaneous human speech. Subsequent research shifted toward \emph{grounded language understanding}, where language is explicitly tied to a robot’s perception and motion. Seminal works\cite{tellex2011understanding,kollar2010toward} introduced probabilistic models that map linguistic elements—like verbs and object references—to real-world entities and spatial relations. This enabled robots to interpret instructions such as “Place the cup on the table” by identifying and manipulating the correct objects and locations. In parallel, deep learning–based approaches have expanded the scope of language instruction following by shifting from rigid grammars to data-driven end-to-end strategies that learn directly from large corpora. Benchmarks and models integrating both visual and linguistic inputs have been introduced for a variety of household tasks, demonstrating the growing synergy between perception and language processing \cite{shridhar2020alfred, jaafar2024lanmp}. More recently, the emergence of large-scale pretrained \emph{foundation models} has allowed robots to interpret and solve complex natural language directives without any training \cite{quartey2024verifiably,paulius2024bootstrapping,liu2024lang2ltl}. \looseness=-1

Despite these advances, a critical gap remains in the use of prosodic features present in speech—such as stress, rhythm, and intonation—to enhance instruction comprehension. Current systems predominantly rely on text transcripts, discarding the acoustic cues that humans naturally employ to convey emphasis and resolve ambiguities. We address this gap by explicitly integrating prosodic cues into the instruction-following pipeline. Our approach enables robots to more effectively interpret emphasis, disambiguate intent, and achieve a robust understanding of tonally ambiguous instructions. \looseness=-1


\begin{figure*}[htbp]
  \centering
  \begin{subfigure}[t]{0.48\textwidth}
    \centering
    \includegraphics[width=\textwidth]{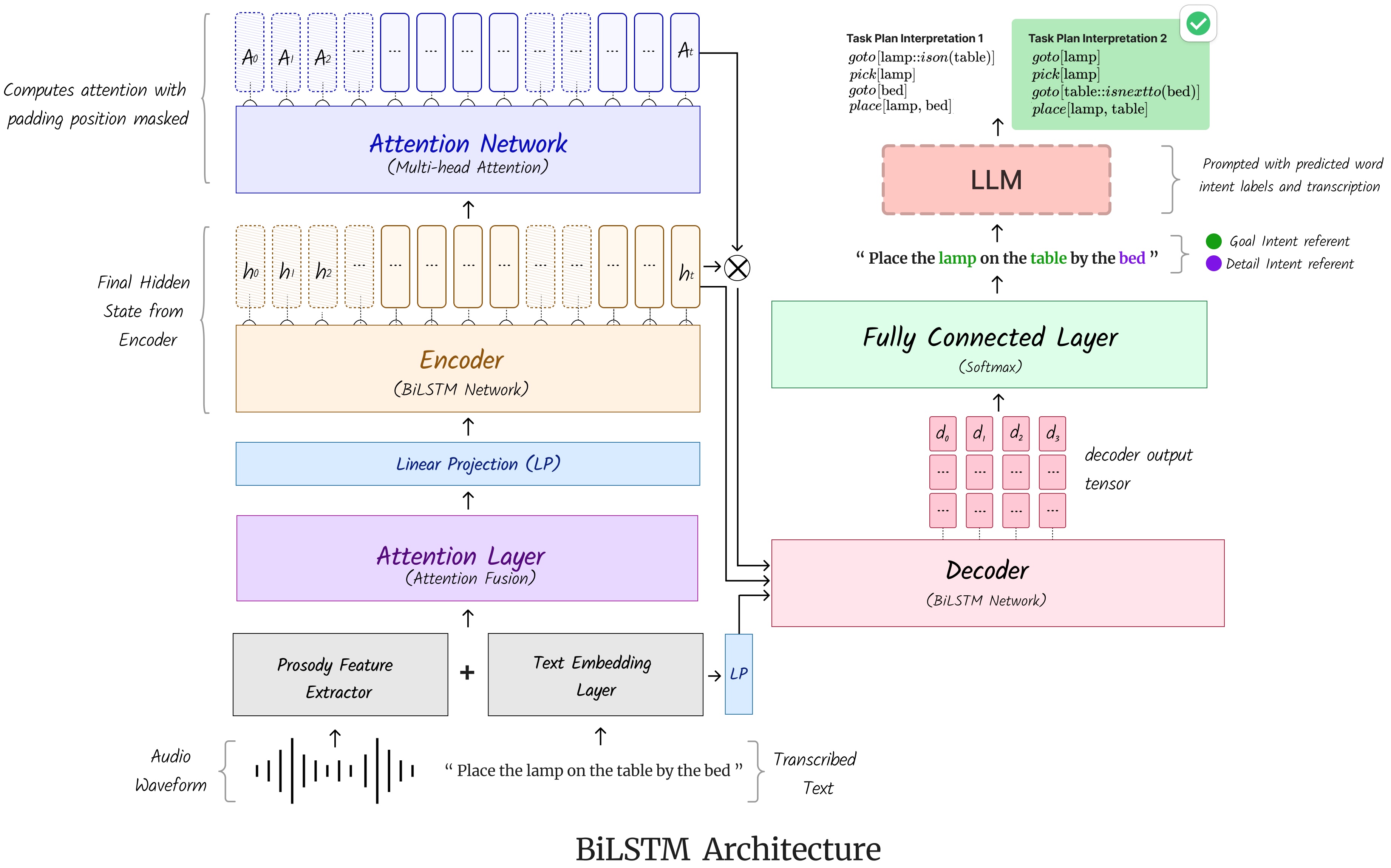}
    \caption{BiLSTM based Encoder-Decoder Model}
    \label{fig:lstm}
  \end{subfigure}
  \hfill
  \begin{subfigure}[t]{0.48\textwidth}
    \centering
    \includegraphics[width=\textwidth]{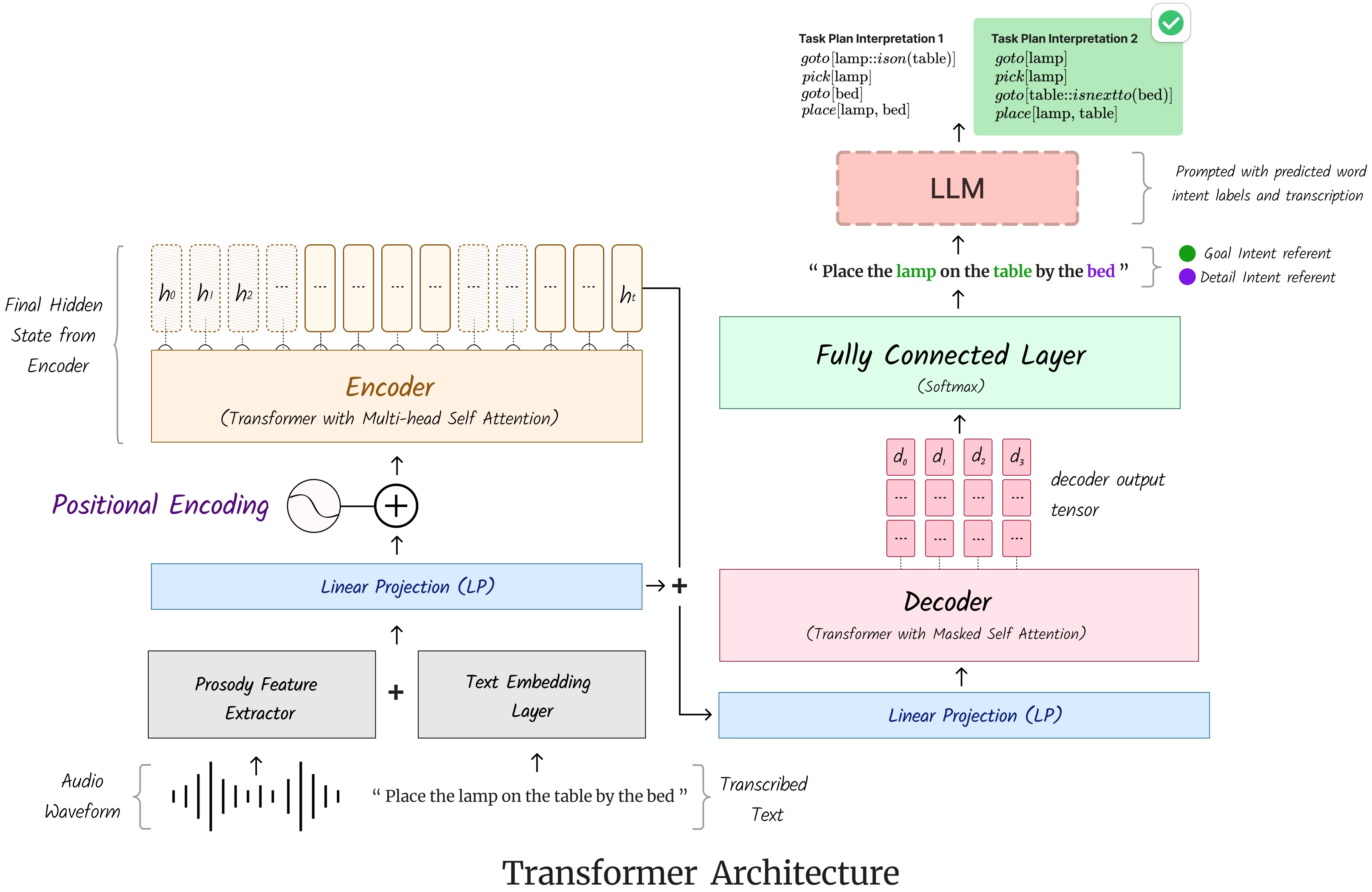}
    \caption{Transformer-based Encoder-Decoder Model}
    \label{fig:transformer}
  \end{subfigure}
  \caption{
Overview of our architectures for word-level intent classification. 
\textbf{(a)~BiLSTM-based Model:} Prosodic features are fused with text embeddings and fed into a BiLSTM encoder. An attention layer refines the fused representations before a decoder predicts intent labels for each token. 
\textbf{(b)~Transformer-based Model:} Prosodic and textual inputs are combined and passed through a Transformer encoder with multi-head self-attention and positional encodings. 
A masked self-attention decoder then generates token-level intents. 
In both models, predicted intent labels are passed to a large language model (LLM) for final task plan selection.  \looseness=-1
}
  \label{fig:combined_models}
      \vspace{-0.4cm}
\end{figure*}

\section{Speech instruction understanding and disambiguation}
We introduce Bidirectional Long Short Term memory (BiLSTM) \cite{zhang-etal-2015-bidirectional} and Transformer-based \cite{Vaswani2017} encoder-decoder models designed to identify referent intents by leveraging prosody. Both architectures are optimized to capture long-range dependencies in prosodic features extracted from speech, enabling the disambiguation of \textbf{goal} and \textbf{detail} intents within a given utterance. \looseness=-1

\subsection{Overview}
\begin{singlespace}
For each speech sample in our dataset, we first perform \textbf{forced alignment} to determine the temporal boundaries of each word. Next, we extract \textbf{prosodic features} for each word using the \texttt{Disvoice} Python library \footnote{Link to Disvoice library: \url{https://disvoice.readthedocs.io/en/latest/index.html}}. Simultaneously, we generate a \textbf{text embedding} using OpenAI's \texttt{Text Embedding Large 3} model \footnote{Link to OpenAi's Text Embedding Large 3 model: \url{https://platform.openai.com/docs/guides/embeddings}}. The extracted prosodic and textual features are then concatenated to form a unified feature representation. This integration is crucial as it allows the model to utilize both the semantic content of words and prosodic cues that can influence meaning and interpretation. \looseness=-1

Formally, we represent the input sequence for the $i$-th speech utterance as:

\begin{equation}
    \mathbf{X^{(i)}= \{x_1^{(i)},x_2^{(i)},...,x_{T_i}^{(i)}\}}
\end{equation}

where each $x_t^{(i)}$ is the feature vector corresponding to the $t$-th word in the utterance, defined as:

\begin{equation}
    x_t^{(i)} \in \mathbb{R}^{d_{\text{prosody}} + d_{\text{embed}}}
\end{equation}

$d_{\text{prosody}}$ and $d_{\text{embed}}$ denote the dimensionalities of the prosodic and textual embedding feature spaces, respectively. These constructed feature vectors are passed to our models, which generate a corresponding sequence of labels:

\begin{equation}
    \mathbf{Y^{(i)}= [y_1^{(i)},y_2^{(i)},...,y_{T_i}^{(i)}]}
\end{equation}

where each $y_t^{(i)}$ represents the predicted label for the $t$-th word in the utterance. Each label indicates whether the word is \textit{not of interest}, a \textbf{goal intent referent}, or a \textbf{detail intent referent}. 

\end{singlespace} 

\subsection{BiLSTM-based encoder-decoder architecture}
In our BiLSTM-based architecture, input prosodic and text embedding features are first processed through an \emph{Attention Fusion} layer, which dynamically assigns importance to different words in the utterance. Each word’s feature vector undergoes a linear transformation followed by a \texttt{tanh} activation to compute an attention score. The computed attention scores are normalized using a softmax function and subsequently applied to their corresponding feature vectors via element-wise multiplication. This results in a fused feature representation that preserves the sequence structure while enhancing the prominence of salient words and suppressing less informative ones. By selectively weighting the input features, the model effectively prioritizes contextually relevant patterns, improving its ability to capture fine-grained variations in speech.

The fused features are passed to the encoder, which begins with a linear projection followed by a \texttt{ReLU} activation to reduce the dimensionality of the input. The output of the linear projection is then fed into a stack of BiLSTM layers, which encode rich contextual representations by capturing both forward and backward dependencies within the sequence. However, since BiLSTM layers struggle to model long-range dependencies effectively \cite{Li2019WhyAA, Vaswani2017}, we incorporate a \emph{Multi-Head Attention} layer to enhance the model’s ability to capture relationships between distant elements in the sequence. The attention scores computed from the \emph{Multi-Head Attention} layer are then applied to the final hidden states of the BiLSTM layers via element-wise multiplication. We concatenate the resulting vectors from this computation with the original word embeddings of the utterance to obtain a final representation, \( z_t \), that encapsulates both prosodic and textual information, ensuring that the decoder receives semantically and acoustically rich features. In the decoder, a stack of BiLSTM layers and a final fully connected layer process these features and assign labels to each word in the input utterance.  \looseness=-1

\subsection{Transformer-based encoder-decoder architecture}
Transformers inherently lack temporal inductive biases \cite{Vaswani2017}, thus in our Transformer-based model we first apply a linear transformation to our input prosodic and text embedding features then integrate fixed positional encodings into input features to provide word order information. This sequence is then processed through a stack of Transformer encoder layers. Each layer employs multi-head self-attention mechanisms to capture global dependencies and relationships between words, irrespective of their positional distances. The depth of the encoder—determined by the number of layers—enables the model to progressively learn abstract representations of the input data. Before decoding, we concatenate the encoder outputs with the initially projected input features and apply another linear projection. This process ensures that the decoder benefits from the detailed acoustic and semantic features present in the original input and the global contextual information encoded by the Transformer.

The decoder generates per-word labels in a left-to-right fashion, consistent with the sequential nature of the prediction task. During training, each target word’s ground truth label is shifted by one position and prefixed with a special start-of-sequence (SOS) token, which serves as the initial input to the decoder. Reintroducing the original embeddings into the decoder improves the model's ability to capture specific word properties during label generation. The decoder uses masked self-attention to maintain the autoregressive property required for sequential prediction. This causal masking ensures that each output position can only attend to preceding positions, preventing information leakage from future tokens. Finally, a fully connected layer with softmax activation maps decoder outputs to a probability distribution over all possible class labels for each word in the sequence. \looseness=-1

\begin{table*}[t]
  \centering
  \small
  \setlength{\tabcolsep}{2pt} 
  \caption{Performance in Classifying Goal and Detail Referent Intents for Transformer and BiLSTM Models.}
  \label{tab:combined}
    \begin{tabular}{ll*{12}{c}}
      \toprule
      & & \multicolumn{4}{c}{Prosody} & \multicolumn{4}{c}{Raw} & \multicolumn{4}{c}{Prosody+Raw} \\
      \cmidrule(lr){3-6} \cmidrule(lr){7-10} \cmidrule(lr){11-14}
      \textbf{Model} & \textbf{Intent} & \textbf{Acc} & \textbf{Prec} & \textbf{Rec} & \textbf{F1} & \textbf{Acc} & \textbf{Prec} & \textbf{Rec} & \textbf{F1} & \textbf{Acc} & \textbf{Prec} & \textbf{Rec} & \textbf{F1} \\
      \midrule
      \multirow{2}{*}{Transformer} 
        & Goal   & \textbf{92.61\%} & \textbf{85.41\%} & \textbf{75.60\%} & \textbf{80.20\%} & 90.81\% & 76.79\% & 76.79\% & 76.79\% & 90.96\% & 84.50\% & 66.51\% & 74.43\% \\
        & Detail & \textbf{94.70\%} & \textbf{83.85\%} & \textbf{81.82\%} & \textbf{82.82\%} & 91.10\% & 80.08\% & 57.27\% & 66.78\% & 89.11\% & 70.16\% & 52.73\% & 60.21\% \\
        & Overall & 93.31\% & \textbf{90.04\%} & \textbf{90.25\%} & \textbf{90.06\%} & \textbf{93.92\%} & 88.13\% & 88.73\% & 88.07\% & 88.59\% & 83.22\% & 83.90\% & 82.95\% \\
      \midrule
      \multirow{2}{*}{BiLSTM} 
        & Goal   & \textbf{93.04\%} & \textbf{97.21\%} & 66.75\%      & \textbf{79.15\%} & 91.48\% & 84.00\% & \textbf{70.33\%} & 76.56\% & 91.90\% & 84.94\% & 67.46\% & 75.20\% \\
        & Detail & \textbf{94.22\%} & \textbf{76.13\%} & \textbf{91.82\%} & \textbf{83.24\%} & 91.43\% & 73.95\% & 69.70\% & 71.76\% & 91.81\% & 72.89\% & 75.76\% & 74.29\% \\
        & Overall & \textbf{95.79\%} & \textbf{92.87\%} & \textbf{92.14\%} & \textbf{91.80\%} & 94.31\% & 88.89\% & 89.39\% & 88.95\% & 94.74\% & 89.50\% & 89.77\% & 89.39\% \\
      \bottomrule
    \end{tabular}
   \vspace{-0.25cm}
\end{table*}

\subsection{Large language modelling head}
We incorporate the per-word goal and detail intent predictions from our speech understanding models into a Large Language Model (LLM) to determine appropriate task plans a robot may follow to satisfy the given instruction. Specifically, we feed the transcribed text utterance along with the predicted intents and in-context examples into an LLM prompt, enabling the model to infer the most appropriate interpretation and subsequent task plan. We leverage the prompting structure and Composable Referent Descriptor (CRD) syntax proposed by \cite{quartey2024verifiably}. To evaluate the effectiveness of different LLMs in this context, we experiment with three high-performing language models from OpenAI: GPT-4o \footnote{\url{https://openai.com/index/hello-gpt-4o/}}, o1-mini\footnote{\url{https://openai.com/o1/}}, and o3-mini\footnote{\url{https://openai.com/index/openai-o3-mini/}}. \looseness=-1

\begin{table}[!htpb]
  \centering
  \small
  \setlength{\tabcolsep}{3pt} 
  \caption{LLM Model Performance in Choosing the Correct Task Plans.}
  \label{tab:llm-per}
  \begin{tabular}{lccc}
    \toprule
           & \textbf{LLM+ASR} & \textbf{\shortstack{LLM+Prosody-\\BiLSTM}} & \textbf{\shortstack{LLM+Prosody-\\Transformer}} \\
    \midrule
    Gpt-4o & 50\%  & \textbf{56.84\%} & \textbf{71.96\%} \\
    o1-mini & 50\%  & 52.10\% & 62.5\% \\
    o3-mini & 50\%  & 54.54\% & 61.83\% \\
    \bottomrule
  \end{tabular}
  \vspace{-0.5cm}
\end{table}
\section{Experimental setup}
\subsection{Data}
To evaluate our approach, we collected a novel accented speech dataset of \textbf{35 ambiguous instructions}, inspired by \cite{quartey2024verifiably}. Each instruction in this dataset can be interpreted in two distinct ways, resulting in \textbf{two seperate sets of intent referents} per instruction. Notably, the goal and detail referents vary between interpretations, introducing structured ambiguity that our models aim to resolve. To create the speech dataset, we recruited \textbf{22 participants} (10 males, 12 females) to record all 35 instructions, providing both possible interpretations. Each participant provided \textbf{44 samples} recordings resulting in \textbf{1,540 voice samples} with a total duration of approximately \textbf{121 minutes}. The recordings were collected using a Google Pixel 8 smartphone. The training set consisted of 1,056 samples, whereas the validation and test sets contained 264 and 220 samples respectively. To encourage generalization, we designed the dataset split such that the instructions present in the evaluation sets were not included in the training data. In our experiments, we conduct feature ablation studies to examine the impact of different input representations on intent detection performance beyond prosodic features alone. In one setting, we extract only raw audio features using the Librosa library \footnote{\url{https://github.com/librosa/librosa}}. In another, we integrate both prosodic and raw audio features as input to our models. This experimental setup allows us to systematically evaluate whether prosodic features alone can effectively enhance the detection of goal and detail intents or if a combination of prosodic and raw features leads to improved performance.\looseness=-1

\subsection{Training procedure}
We train our models by minimizing the \textit{cross-entropy loss} between the predicted label sequences, $\hat{Y}^{(i)}$, and the corresponding ground truth, $Y^{(i)}$. To identify the optimal hyperparameter configurations, we conduct a \textit{grid search}, using the Optuna hyperparameter optimization framework \footnote{\url{https://optuna.org}}, exploring a range of values for the learning rate, hidden dimension, number of layers, dropout rate, attention-layer depth, and weight decay. The values that maximize the \textit{F1-score} on the validation set are selected. The \textit{optimal learning rate} for the Transformer-based model is found to be $2.22 \times 10^{-4}$, while the BiLSTM-based model performs best with a learning rate of $4.2 \times 10^{-3}$. The \textit{hidden dimension} of the linear layers is optimized at $448$ for the Transformer-based model and $512$ for the BiLSTM-based model. Regarding model depth, the \textit{best-performing Transformer model} consists of \textit{three layers}, whereas the \textit{optimal BiLSTM model} is a \textit{single-layer architecture}. For \textit{regularization}, the dropout rate is optimized at $0.25$ for the Transformer model and $0.45$ for the BiLSTM model. The Transformer model's attention mechanism benefits from a deeper configuration, consisting of \textit{eight attention layers}, whereas the BiLSTM model operates effectively with a \textit{single attention layer}. Finally, weight decay, which plays a crucial role in controlling overfitting, is optimized at $2.25 \times 10^{-6}$ for the Transformer-based model and $5.44 \times 10^{-5}$ for the BiLSTM-based model. The total number of parameters for our BiLSTM-based model is \textit{18,153,476} while the total number of parameters for our Transformer-based model is \textit{19,003,014}. During training, we optimize model parameters using the Adam \cite{Kingma2014AdamAM} optimizer with weight decay. A \textit{learning-rate scheduler} is employed to dynamically adjust the step size if progress stagnates. It takes on average 8 hours to perform a hyperparameter search and train our models.  \looseness=-1

\section{Results}

\subsection{Referent detection task performance}
Our experimental results indicate that utilizing \textbf{prosodic features} alone consistently yields the highest performance in goal and detail referent detection tasks. These findings suggest that while raw acoustic features are informative, \textbf{prosodic signals exhibit superior discriminative power}, effectively highlighting key acoustic patterns associated with instruction intent that may otherwise be obscured within raw audio signals. Architecturally, the \textbf{BiLSTM model slightly outperforms the Transformer model} in referent detection, which aligns with expectations, as Transformer models are known to benefit more significantly from larger training datasets \cite{niu2024beyond}.

\subsection{Intent Disambiguation task performance} 
For the intent disambiguation task, which involves selecting the appropriate task plan given a speech utterance, our experimental results, presented in Table \ref{tab:llm-per}, demonstrate that augmenting LLM input prompts with goal and detail intent referents detected by our proposed prosody-based models significantly enhances performance. Notably, the best performance on this task is achieved by the combination of our Prosody-Transformer model and GPT-4o. We hypothesize that GPT-4o outperforms the other LLMs due to its likely larger model size and richer internal representations, which enable superior generalization across diverse tasks.

\section{Limitations and conclusion}
Our work is limited by a relatively small dataset (1540 samples) and the narrow age range (18--22) of participants, potentially restricting the generalizability of our findings. Our future efforts will focus expanding the dataset size and participant diversity to enhance model robustness and applicability across varied speaker populations. In this study, we have demonstrated the pivotal role of prosodic information in accurately interpreting speech instructions for human-robot interaction. By incorporating prosodic cues, robots can more effectively infer intent, thereby improving the naturalness of human-robot communication. We also introduce speech disambiguation as a key research area in robotics and present a first-of-its-kind dataset to catalyze further advances in this field. \looseness=-1


\end{document}